\documentclass[10pt,twocolumn,letterpaper]{article}

\usepackage{iccv}
\usepackage{times}
\usepackage{epsfig}
\usepackage{graphicx}
\usepackage{amsmath}
\usepackage{amssymb}
\usepackage{float}
\usepackage{caption}
\usepackage{subcaption} 
\usepackage{mathrsfs}
\usepackage{booktabs}
\usepackage{gensymb}
\usepackage{esvect}
\usepackage{url}

\usepackage{iccv}
\usepackage{times}
\usepackage{epsfig}
\usepackage{graphicx}
\usepackage{amsmath}
\usepackage{amssymb}

\usepackage[breaklinks=true,bookmarks=false]{hyperref}

\iccvfinalcopy 

\def\iccvPaperID{****} 

\ificcvfinal\fi

\begin{document}

\title{End-to-end Hand Mesh Recovery from a Monocular RGB Image}

\author{Xiong Zhang$^{1* }$, Qiang Li$^{1*}$, Hong Mo$^2$, Wenbo Zhang$^{1\dagger}$, Wen Zheng$^1$\\
$^1$Y-tech, Kwai, $^2$State Key Laboratory of VR, Beihang University\\
{\tt\small $^1$\{zhangxiong,liqiang03,zhangwenbo,zhengwen\}@kuaishou.com, $^2$mandymo@buaa.edu.cn}
}


\twocolumn[{%
\renewcommand\twocolumn[1][]{#1}%
\maketitle
\begin{center}
	\centering
	\includegraphics[width=\textwidth]{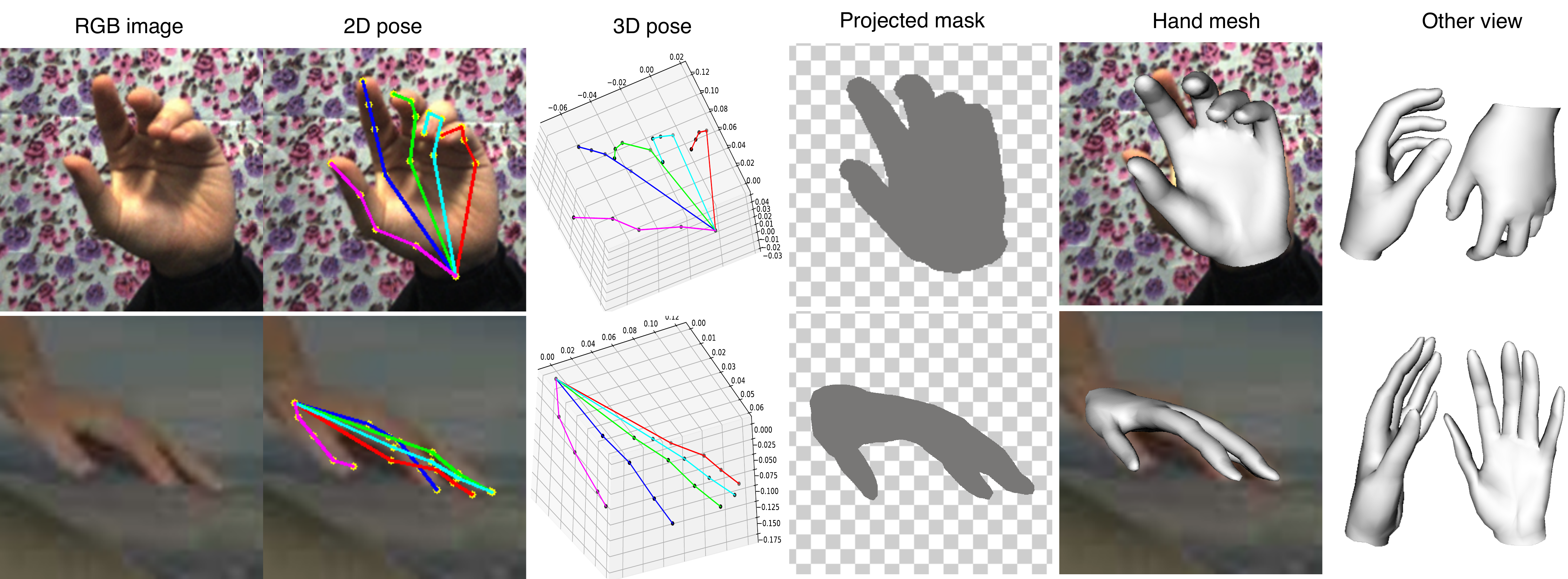}
	\label{effect_title_image}
\captionof{figure}{\textbf{Hand Mesh Recovery.}
Our framework provides a unified scheme for monocular hand image understanding.
The first column shows two examples from the STB \cite{hand_stb} and RHD \cite{hand_zb} dataset respectively.
The second and third columns present the estimated 2D joints and 3D joints locations.
The fourth column demonstrate the projection mask of the reconstructed hand mesh,
while the last two columns illustrate the recovered mesh from several viewpoints.
Note that our framework can recover the mesh in a reasonable way even in the presence of severe occlusions.}
\label{effect_title_image}
\end{center}%

}]

\thispagestyle{empty}


\begin{abstract}
In this paper, we present a HAnd Mesh Recovery (HAMR) framework to tackle the problem of reconstructing the full 3D mesh of a human hand from a single RGB image.\footnote{
The code is available at {https://github.com/MandyMo/HAMR}.\\
$~~\quad$*Indicates equal contribution.\\
$~~\quad\dagger$Corresponding author.}
In contrast to existing research on 2D or 3D hand pose estimation from RGB or/and depth image data, HAMR can provide a more expressive and useful mesh representation for monocular hand image understanding.
In particular, the mesh representation is achieved by parameterizing a generic 3D hand model with shape and relative 3D joint angles.
By utilizing this mesh representation, we can easily compute the 3D joint locations via linear interpolations between the vertexes of the mesh, while obtain the 2D joint locations with a projection of the 3D joints.
To this end, a differentiable re-projection loss can be defined in terms of the derived representations and the ground-truth labels, thus making our framework end-to-end trainable.
Qualitative experiments show that our framework is capable of recovering appealing 3D hand mesh even in the presence of severe occlusions.
Quantitatively, our approach also outperforms the state-of-the-art methods for both 2D and 3D hand pose estimation from a monocular RGB image on several benchmark datasets.
\end{abstract}

\vspace{-2ex}
\section{Introduction}

Hand image understanding targets the problem of recovering the spatial configuration of hands from natural RGB or/and depth images, which has many applications such as human-machine interaction \cite{markussen2014vulture, sridhar2015investigating}, and virtual/augmented reality \cite{lee2009multithreaded,piumsomboon2013user}.
Estimating the spatial configuration is very challenging due to the diverse appearance variation, self occlusions and complex articulations.
While many existing works considered markerless image-based hand image understanding,
most of them require depth cameras \cite{baek2018augmented,garcia2018first,moon2018v2v,qian2014realtime,sridhar2015fast,tagliasacchi2015robust,taylor2016efficient,yuan2018depth}
or multi-view images \cite{ballan2012motion,gomez2017large,simon2017hand,sridhar2013interactive,wang20116d} to handle the above difficulties.
As a result, most of these methods are impractical for real-world scenarios where only monocular RGB images are available.

For the monocular RGB setting, many recent works have investigated the specific problems of 2D and 3D hand pose estimation \cite{hand_geliuhao,iqbal2018hand,hand_muller,simon2017hand,spurr2018cross,hand_zb}.
Considering 2D hand pose,
Simon \emph{et al.} \cite{simon2017hand} utilized bootstrapping to estimate 2D joint locations from single-view RGB images, while the approach has to leverage multi-view images for 3D cases.
Gomez-Donoso \emph{et al.} \cite{gomez2017large} introduced a simple yet effective deep learning architecture for 2D hand pose estimation and also built a multi-view 3D hand pose dataset.
Note that single RGB image based 2D hand pose estimation is prone to unreasonable predictions under self occlusions and complex configurations.

Regarding 3D hand pose estimation from monocular RGB,
an obvious difficulty resides in the perspective ambiguity caused by missing depth information.
Zimmermann and Brox \cite{hand_zb} presented the first learning based formulation which augments deep networks to learn a network-implicit 3D articulation prior for resolving perspective ambiguities.
Panteleris \emph{et al.} \cite{panteleris2018using} formulated the 3D pose estimation from 2D pose as an inverse kinematics (IK) problem which constrains the solution space to only plausible hand articulations.
Mueller \emph{et al.} \cite{hand_muller} also introduced several geometry-consistent rules to enforce human hand constraints to relief the hand parsing ambiguity.

Besides enforcing geometric constraints, several recent works have investigated intricate learning strategies.
Spurr \emph{et al.} \cite{spurr2018cross} designed a generative hand model represented by a coherent latent space across RGB, 2D and 3D pose modalities, which enables semi-supervised learning.
Iqbal \emph{et al.} \cite{iqbal2018hand} introduced an intermediate 2.5D pose representation which can be easily estimated from RGB images, and then provided a solution to reconstruct the 3D pose from 2.5D.
Cai \emph{et al.} \cite{hand_geliuhao} proposed a weakly-supervised method for 3D pose regression by leveraging depth images during training.
Though effective at large, the two types of strategies are insufficient to fully capture the inherent properties and diversity of hand articulations.

Aside from the perspective ambiguity, another challenge is the lack of high-quality annotated 3D hand pose data.
To address this issue, Zimmermann \emph{et al.} \cite{hand_zb} turned to render synthetic data with 3D models, from which the ground truth annotations of 3D joints can be easily obtained.
Mueller \emph{et al.} \cite{hand_muller} employed CycleGAN \cite{zhu2017unpaired} to synthesize training data which can enhance the performance of 3D hand pose estimation to some extend.
However, networks trained on synthetic data do not generalize well to real-world images due to the domain shift.

In this paper, we propose a hand mesh recovery (HAMR) framework which is also able to produce 2D and 3D pose estimation as side products, as shown in Fig. \ref{effect_title_image}.
In contrast to current research on 2D or 3D hand pose estimation, HAMR provides a more comprehensive and flexible mesh representation for hand RGB image understanding.
Inside its core, HAMR parameterizes a generic 3D hand model to define the mesh representation in terms of shape and relative 3D joint angles.
With this mesh representation, the 3D joint locations can be computed by linearly interpolating between mesh vertexes.
In addition, one can obtain the 2D joint locations with a simple projection of the 3D joints.

As a result, HAMR provides a graceful approach to handle perspective ambiguities and data insufficiency.
Firstly, the introduced parametric model in HAMR has implicitly characterized the inherent properties of hand comprehensively.
This is also validated by our ablation study that adding several specific geometric constraints over fingers can yield very slight improvements.
Secondly, since HAMR is end-to-end trainable thanks to the differentiable re-projection loss, our method can exploit enormous incompletely annotated 2D and 3D training samples.
The effectiveness of HAMR is evaluated via extensive qualitative and quantitative experiments on benchmark datasets.

\begin{figure*}[ht]
\centering
\includegraphics[width=0.95\textwidth]{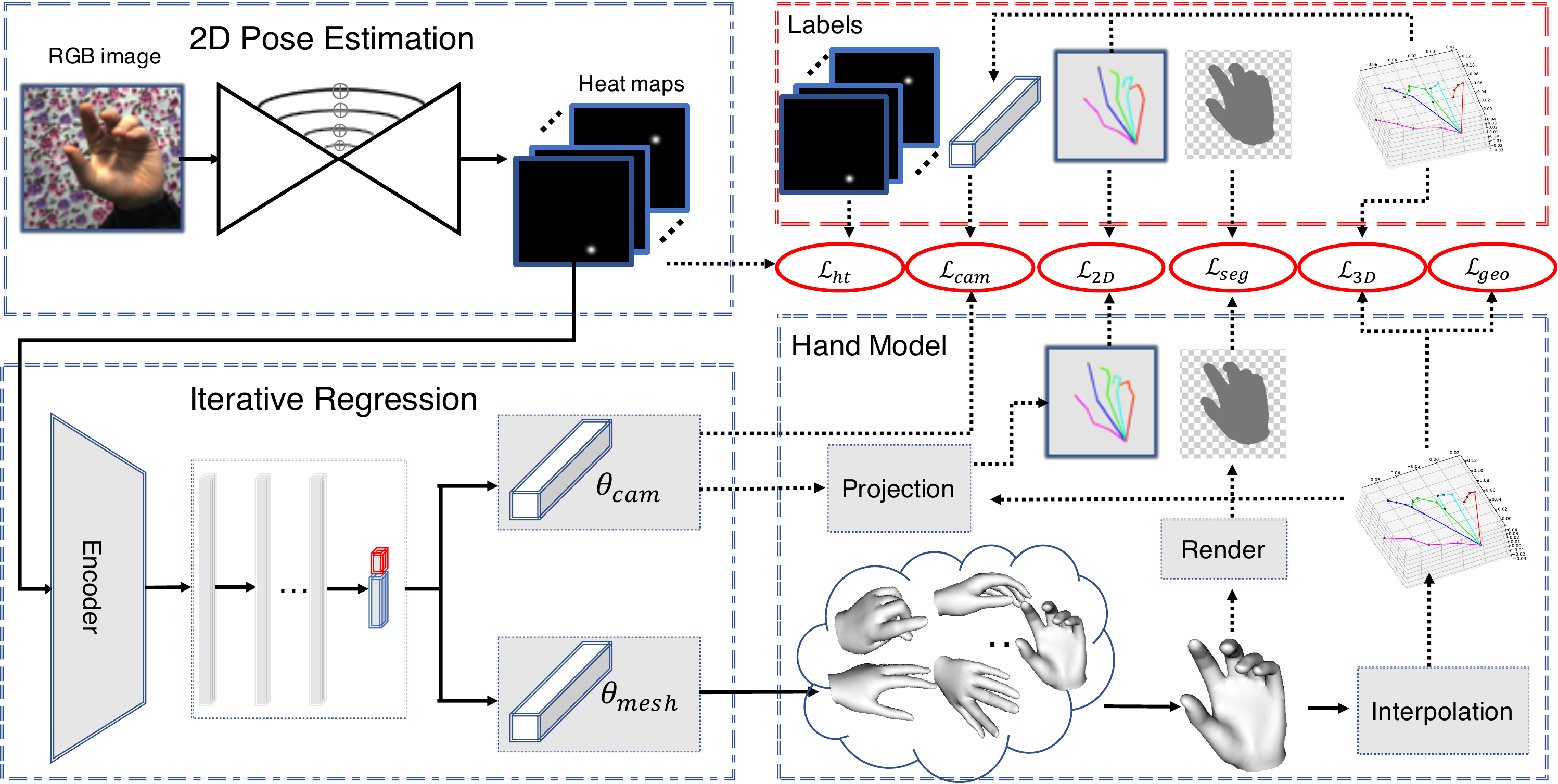}
\caption{{\bf Framework Architecture}. Our framework consist of three parts: (1) a 2D pose estimation block aims at estimating 2D heat-maps from a single RGB image, (2) a regression module that regresses for 3D parameters $\Theta=\{\theta_{mesh}, \theta_{cam}\}$ by a iterative way, (3) a mesh recovery part for generating the mesh representation, the 2D and 3D joints locations from $\Theta$.
}
\label{main_architecture_framework}
\end{figure*}


\section{Related Work}
In literature, 3D reconstruction is a long standing problem that has received considerable attention in computer vision and computer graphics.
Here, we present a brief overview of two particular topics on 3D face and human body reconstruction from a single RGB image.

{\bf Monocular Face Reconstruction.} Thanks to the 3D morphable model (3DMM) released by \cite{blanz1999morphable,cao2014facewarehouse,paysan20093d}, the 3D face reconstruction from single RGB has achieved great success in recent yeas.
For example, MoFA \cite{tewari2017mofa} provided an end-to-end trainable framework by leveraging 3DMM to define a differential parametric decoder for face pose, shape, expression, skin reflectance and scene illumination.
As a step forward, Tewari {et al.} \cite{tewari2018self} further considered a regressor for the face properties in addition to the differential parametric model.
Genova \emph{et al.} \cite{genova2018unsupervised} proposed a method for training a regression network for fitting parameters of 3DMM using only unlabeled photographs.
Sengupta \emph{et al.} \cite{sengupta2018sfsnet} implemented an end-to-end learning framework for producing an decomposition of an unconstrained human face image into shape, reflectance and illuminance.

{\bf Monocular Human Body Recovery.} Similar progress have been achieved on human body mesh recovery from single RGB.
For instance, Bogo \emph{et al.} \cite{Bogo:ECCV:2016} presented the first method to automatically estimate the 3D pose of the human body as well as its 3D shape from a single unconstrained image.
Kanazawa  \emph{et al.} \cite{Kanazawa:CVPR:2018} described an end-to-end framework for recovering a full 3D mesh of a human body from a single RGB image by employ a CNN to regression of the parameters of a parametric model SMPL \cite{SMPL:2015}. 
Similar to \cite{Kanazawa:CVPR:2018}, Omran \emph{et al.} \cite{NBF:3DV:2018} performed a novel network structure for human mesh recovery and achieve state-of-the-art result on task of 3D human pose estimation.
Recently, BodyNet \cite{Varol_2018_ECCV} provided an end-to-end trainable network that considered a volumetric 3D loss, a multi-view re-projection loss, and several kinds of intermediate supervision.

Aside from the above researche works, hand reconstruction from monocular RGB has not drawn sufficient attention from the community. To the best of our knowledge, HAMR provides one of the first approaches to achieve this target
among other concurrent works \cite{boukhayma20193d, ge20193d}.
\vspace{-0.5ex}

\section{Framework}
\label{sec:Framework}
The main goal of this paper is to present a unified framework for providing a detailed mesh representation $\mathcal{M} \in \mathbb{R}^{N \times 3}$ together with 2D key points $\Phi^{2D} \in \mathbb{R}^{K \times 2}$ and 3D joint locations $\Phi^{3D} \in \mathbb{R}^{K \times 3}$ of hand from a RGB image centered on hand, where $N$ is the vertex count of $\mathcal{M}$ and $K$ indicate the number of key points. 
We exploit a recent parametric generative model for generating hand mesh, then infer hand pose from the generated mesh.
The overall architecture of our framework is illustrated as Fig.~\ref{main_architecture_framework}.

\subsection{Hand Mesh Recovery}

\textbf{Hand Mesh Representation.} We utilize MANO \cite{mano} as the generic 3D hand model in our framework.
In particular, MANO factors hand mesh into \emph{shape} - mainly models hand properties such as finger slenderness and palm thickness and \emph{pose} - how the 3D surface deforms with articulation. 
Similar to SMPL \cite{SMPL:2015}, MANO parameterizes a triangulated mesh $\mathcal{M} \in \mathbb{R}^{N\times3}$ with a set of parameters $\theta_{mesh}=\{\vv\beta, \vv\theta\}$, where $\vv{\beta} \in \mathbb{R}^{10}$ denote the shape parameters and $\vv\theta \in \mathbb{R}^{K\times3}$ are the pose parameters.
Technically speaking, $\vv{\beta}$ represents the coefficients of PCA components that sculpting the identity subject and $\vv\theta$ denotes the relative 3D rotation of $K$ joints on Rodrigues Vector representation.
The model is defined by a mean template mesh ${\bf \bar{T}} \in \mathbb{R}^{N\times3}$ represented by concatenating $N$ vertices in rest pose $\vv\theta^*$;
a set of blend weights $\mathcal{J} \in \mathbb{R}^{N\times K}$ and $\mathcal{W} \in \mathbb{R}^{N \times K}$;
a blend function $B_S: \mathbb{R}^{|\vv{\beta}|} \mapsto \mathbb{R}^{N \times 3}$ that takes $\vv{\beta}$ as input and outputs a blend shape to characterize the identity subject; 
a pose dependent blend function $B_P: \mathbb{R}^{|\vv{\theta}|} \mapsto \mathbb{R}^{N \times 3}$  that compensates for the deformation caused by pose variations.



Following standard skinning procedure, the shape and pose deformations $B_S(\vv{\beta})$, $B_P(\vv\theta)$ are applied to the mean template $\bf {\bar {T}}$.
To obtain the final mesh, the mesh is then posed by rotating each finger part around joints $J(\vv{\beta})$ using the blend skinning function $W(\cdot)$,
\begin{align} \label{Eq_mesh_rep}
	M(\vv{\beta}, \vv\theta)&=W(T(\vv{\beta},\vv\theta),J(\vv{\beta}),\vv\theta,\mathcal{W}), \\
	T(\vv{\beta},\vv\theta) &= {\bf {\bar {T}}} + B_S(\vv{\beta}) + B_P(\vv\theta).
\end{align}
With the mesh model, one can easily reconstruct a hand mesh by specifying certain values of the parameters $\vv\beta$ and $\vv\theta$.
It is worth mentioning that both $T(\cdot)$ and $W(\cdot)$ are differentiable functions of $\vv\beta$ and $\vv\theta$ \cite{SMPL:2015,mano}, which are favorable for composing the mesh model with other modules.

\textbf{Derived Hand Pose Representations.}
Given the recovered mesh,
we can compute the 3D joint locations $\Phi^{3D}$ via linear interpolations between the vertexes of the mesh,
while obtain the 2D joint locations $\Phi^{2D}$ with a projection of the 3D joints. More specifically,
\begin{align}
\Phi^{3D}_k &= \sum_{i=1}^N \mathcal{J}_{ik}\mathcal{M}_i, \label{es_3d}\\
\Phi^{2D}_k &= \prod\left(\Phi^{3D}_k | \theta_{cam}\right), \label{es_2d}
\end{align}
where $\Phi^{3D}$ is a set of 3D coordinates $\{(x,y,z)_k\}_{k=1}^K$, $\Phi^{2D}$ is a set of corresponding 2D coordinates $\{(u, v)_k\}_{k=1}^K$, and $\theta_{cam}=\{(s,t_x,t_y)\}$ denotes the camera parameter.
The 3D-to-2D projection function $\prod(\cdot)$ is defined as below,
\begin{align}\label{Eq_projection}
	\prod\left((x,y,z)|(s,t_x, t_y)\right) &= \left(s(x+t_x), s(y + t_y)\right),
\end{align}
where we used the weakly perspective camera model.

\textbf{Silhouette Constraints.} By impose supervision over 2D/3D hand pose only, the mesh may simply deform to best fit joint locations and ignoring hand surfaces, which lead to abnormal mesh representation. 
For a hand image, the projection of a well-defined hand mesh should be consistent with the corresponding silhouette. 
This constraint plays an indispensable role in single RGB image based 3D reconstruction \cite{blanz1999morphable, Bogo:ECCV:2016, kato2018neural, pavlakos2018learning, tung2017self}. 
In this work, we introduce the silhouette consistent loss to refine the hand mesh and camera parameter estimation. In particular, we utilize the Inverse Graphics technique \cite{kato2018neural, marschner1998inverse} to obtain a rendered mask image from the reconstructed hand mesh given by Eq. \ref{Eq_mesh_rep}, then penalize misalignment between the rendered mask and the ground-truth silhouette to train the network.

\textbf{Geometric Constraints.}
Upon the derived 3D hand pose representation, it is also helpful to apply several kinds of geometric constraints.
Let $P_a, P_b, P_c, P_d$ be the 4 joints of the fingers (except thumb) in turn (tip to palm order), where $P_i \in \mathbb{R}^3, i \in \{a, b, c, d\}$. 
For convenience, we define $\vv{V_{ab}}$ as $P_a - P_b$, which is similar to $\vv{V_{bc}}$ and $\vv{V_{cd}}$.
A correct hand articulation should obey the following constraints.
First, $P_a, P_b, P_c, P_d$ are in the same plane, which implies:
\begin{equation}
 (\vv{V_{ab}} \times \vv{V_{bc}}) \cdot \vv{V_{cd}} = 0.
\label{eq_geo_1}
\end{equation}
Second, the rotate direction from $\vv{V_{ab}}$ to $\vv{V_{bc}}$ is consistent with that of  $\vv{V_{bc}}$ to  $\vv{V_{cd}}$, which implies:
\begin{equation}
(\vv{V_{ab}} \times \vv{V_{bc}}) \cdot (\vv{V_{bc}} \times \vv{V_{cd}}) \ge 0.
\label{eq_geo_2}
\end{equation}
Note that these two types of geometric constraints can be re-formulated as regularizers in a loss minimization problem by the Lagrangian multiplier method \cite{rockafellar1993lagrange}.

\textbf{Loss Function.}
For our target, to recover the hand mesh from a single RGB image, we leverage deep convolutional networks to fit the mesh parameters $\theta_{mesh}$.
However, in real-world scenario, it is almost impossible to obtain the ground-truth mesh parameters when annotating from single RGB images.
Fortunately, our HAMR framework can define derived 3D and 2D joint locations from the mesh.
By doing so, we can train HAMR with widely-available 3D and 2D annotations thus enable the mesh reconstruction.

We shall point out there is an issue coming from the camera parameters $\theta_{cam}$.
To make our framework fully automatic and accurate, we direct another flow from the base deep convolutional networks to estimate the camera parameters.
During training, the ``ground-truth" camera parameters are computed from the ground-truth paired 3D and 2D annotations.
More specifically, the ``ground-truth" scale $s$ is calculated as the ratio between the average 2D and 3D bone lengths.
Then, the ``ground-truth" offsets are computed as $(t_x,t_y)=\frac1K\sum_{k=1}^{K}((u_k,v_k)/s-(x_k,y_k))$.

Considering training the base deep convolutional networks, we resort to the intermediate supervision to make sure the earlier stage well captures the image semantics. In particular, the 2D pose estimation block is supervised by stacked heatmaps of 2D ground-truth labels. For each 2D joint, the ground-truth heatmap is defined as a Gaussian-blurred confidence image centered at that joint location.

To this end, we employ L2 loss between the derived 3D and 2D representations and ground-truth labels which result in $\mathcal L_{3D}$ and $\mathcal L_{2D}$ respectively.
In addition, the geometric constraints are reformulated as regularizers leading to $\mathcal L_{geo}$ which defines over the predicted 3D poses.
Equally important,  we penalize the misalignment between the rendered mask and ground-truth silhouette via L1 loss leading to $\mathcal L_{seg}$.
Furthermore, L2 loss is utilized to supervise the estimated camera parameters with ground-truth camera parameters leading to $\mathcal L_{cam}$.
As for the intermediate supervision, we compute the pixel-wise distances between the predicted and ground-truth 2D heatmaps that yield $\mathcal L_{ht}$.

The whole process is fully differentiable with respect to all the learnable parameters, thus making our HAMR framework end-to-end trainable.
The overall loss function is summarized as
\begin{equation}\label{total_loss}
\begin{aligned}
	\mathcal{L} &= \lambda_{3D}\mathcal{L}_{3D} + \lambda_{2D}\mathcal{L}_{2D} + \lambda_{geo}\mathcal{L}_{geo}\\
	&\quad + \lambda_{cam} \mathcal{L}_{cam} +  \lambda_{ht}\mathcal{L}_{ht} + \lambda_{seg}\mathcal{L}_{seg},
\end{aligned}
\end{equation}
where $\{\lambda_{3D}, \lambda_{2D}, \lambda_{geo}, \lambda_{cam}, \lambda_{ht}, \lambda_{seg}\}$ are hyper-parameters to trade-off among different types of supervision over the whole framework.

\subsection{Iterative Regression Module}
\label{sec:Framework_Regression_Block}
\begin{figure}
\centering
\includegraphics[width=0.9\columnwidth]{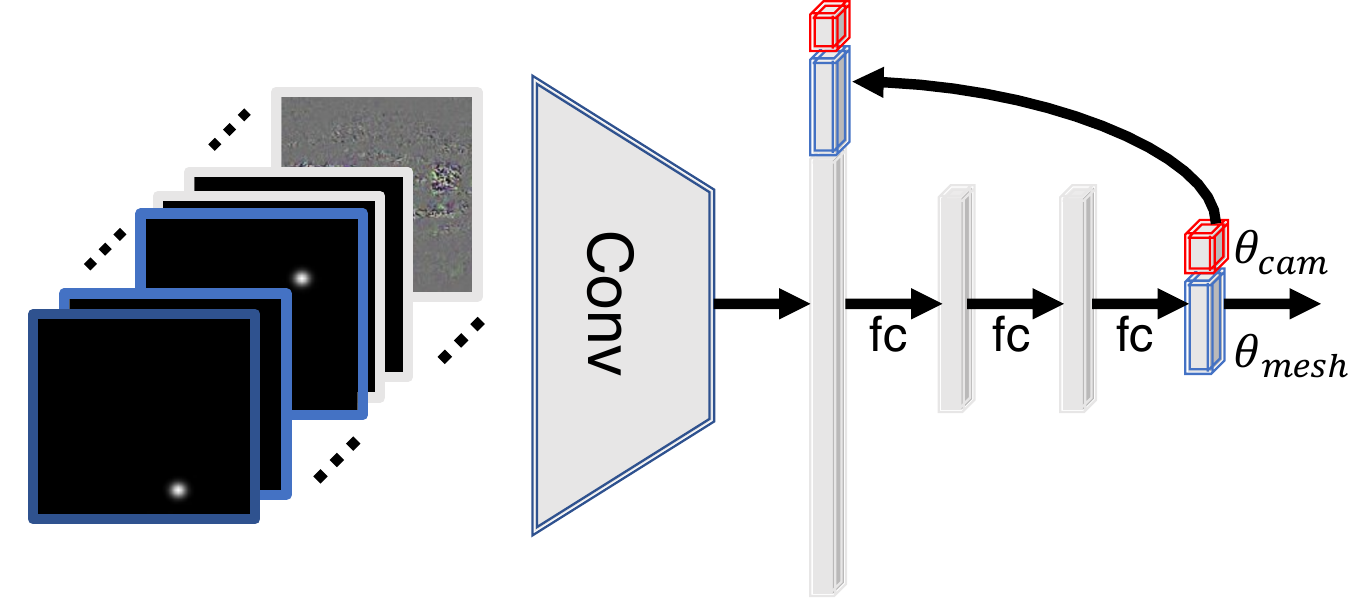}
\caption{{\bf Architecture of Iterative Regression Module}. This module takes the cross-level features as input and regresses the camera and mesh parameters in an iterative way.}
\label{architecture_iterative_regression_block}
\end{figure}
We apply a regression module to fit the camera parameters $\theta_{cam}$ and the mesh parameters $\theta_{mesh}$. However, the complicated domain gap makes it difficult to produce reasonable estimates in one go. Several works \cite{carreira2016human,casecade_pose,oberweger2015training, pose_cpm, casecade_face} revealed that a cascade and coarse-to-fine method shall be more appropriate than a one-pass solution. Inspired by these studies, we implement an iterative regression module to fit the camera and mesh parameters from semantic features extracted from previous 2D pose module. Formally, we take the current parameters ($\theta_{cam}^t,\theta_{mesh}^t$) as additional inputs upon the image feature $\phi$, and estimate the more accurate ($\theta_{cam}^{t+1},\theta_{mesh}^{t+1}$).
As illustrated in Fig. \ref{architecture_iterative_regression_block}, the iterative regression module consists of a simple fully convolutional encoder and multiple fully connected layers.




To make the predicted $\theta_{cam}$ more accurate, we enforce a strong supervision by utilizing the ground-truth camera parameters computed from paired 3D and 2D annotations.
The camera parameters $\theta_{cam}$ consist of three real numbers $(s, t_x, t_y)$ for projecting 3D joint $(x,y,z)$ to 2D key point $(u,v)$,
and the projection function follows Eq. \ref{Eq_projection}.

\subsection{2D Pose Estimation}
\label{sec:Framework_PosetModel}
Similar to recent methods \cite{martinez2017simple,pose_ht_3d_weakly,hand_zb}, we employ a cascade encoder-decoder style network to predict the 2D Gaussian like heat-maps $\Phi^{ht} \in \mathbb{R}^{K\times H\times W}$ for the 2D pose estimation task, where $K$ indicates the number of joints and $\{H, W\}$ are the resolutions of the heat-maps. Each key point has a corresponding heat-map, and each pixel value on the heat-map indicates the confidence of the key point locating in that 2D position. Further, we share the similar view with \cite{hand_geliuhao,hand_muller} that one can hardly resolve perspective ambiguity when applying a direct regression from heat-maps to 3D pose. Actually, different 3D joint locations may be projected to the same 2D key point.
Therefore, aside from the $K$ heat-maps for the 2D key point's locations, this block outputs an extra $M-K (M>K)$ heat-maps to handle the issue.
In addition, we concatenate the intermediate-layer features with the $M$ heat-maps, and feed them into the following iterative regression module.

As for the network architecture, we utilize the Stacked Hourglass network \cite{hourglass} for 2D hand pose estimation with several slight modifications. Similar to \cite{new_hourglass}, we replace the residual block with simple 
$3 \times 3$ convolution, and use average-pooling to replace the max-pooling. Besides, a batch-normalization layer is added after each $3 \times 3$ convolution layer for faster training.

\begin{figure*}
\centering
\includegraphics[width=0.9\textwidth]{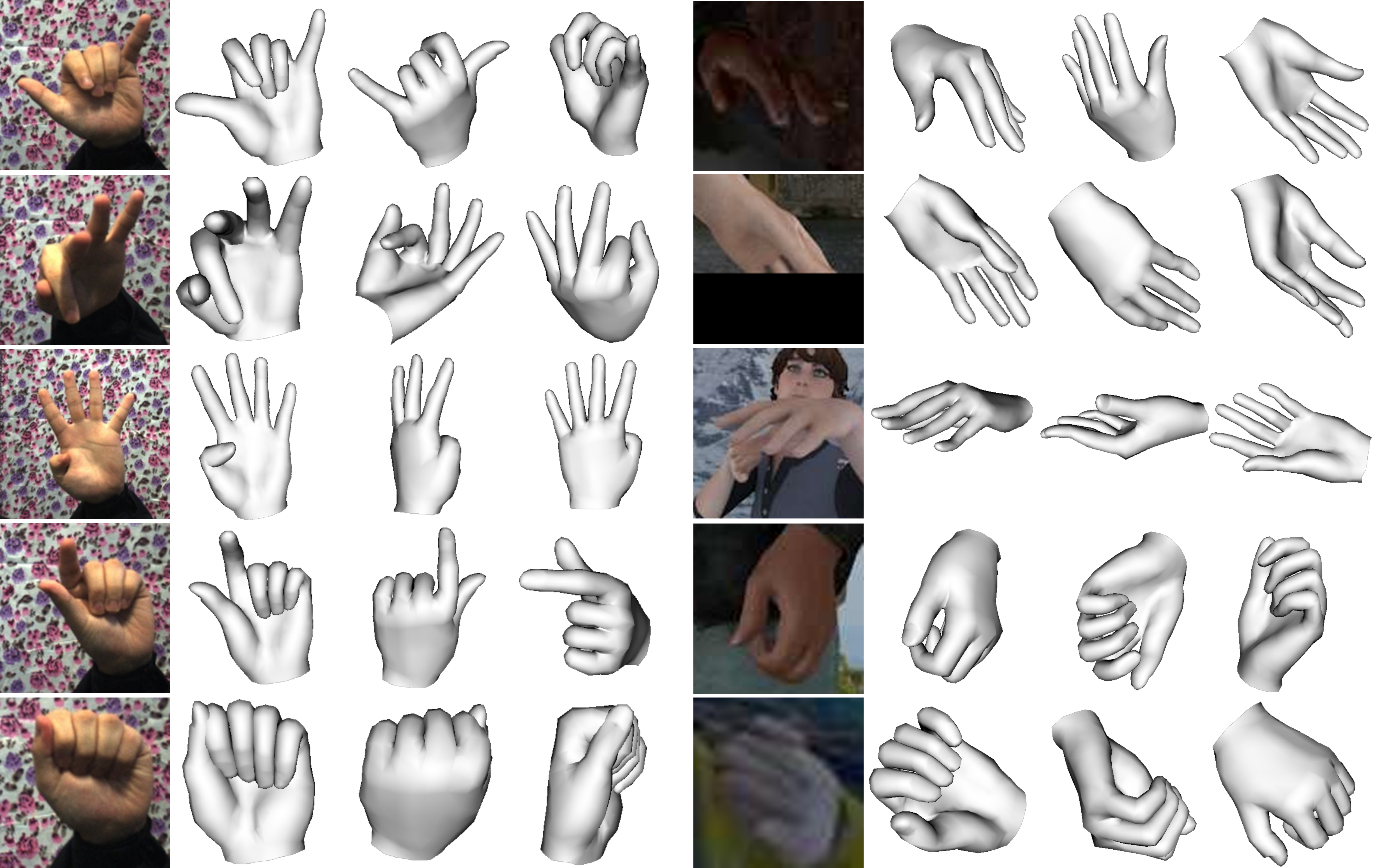}
\caption{{\bf Mesh Recovery Results.} The left four columns present five representative examples from the STB dataset, while the right four columns from the RHD dataset. For each sector, the first column shows the RGB images, the second to fourth columns are the rendered meshes from different views.}
\label{appealing_mesh}
\end{figure*}

\section{Experiments}
To evaluate the efficacy of HAMR, we first present qualitative results on the recovered mesh from single RGB images. 
Secondly, since there exists no ground-truth mesh representation for comparison study, we quantitatively evaluate the superiority of HAMR on the task of 3D and 2D hand pose estimation instead.
Finally, we perform ablation studies to better understand the impact of different design strategies on the 3D hand pose estimation task.\footnote{In practice, the MANO model only has 16 key points and lacks 5 fingertip endpoints. Thus, in our experiments, we manually select vertexes with the index of 734, 333, 443, 555, 678 from generated mesh as the locations of five fingertip endpoints.}
\subsection{Experiment Settings}
{\bf{Datasets.}}
We mainly involve the Rendered Hand Dataset (RHD) \cite{hand_zb}, the Stereo Hand Pose Tracking Benchmark (STB) \cite{hand_stb} and the Dexter Object (Dexter) \cite{hand_dextor} dataset.
The RHD is a 3rd view based synthetic dataset that provides 41258 training and 2728 testing samples, and each sample contains an RGB image, a depth image, a segmentation mask image and both 2D and 3D annotations of the 21 standard key points. In this paper, we follow the standard training/testing division of RHD. Furthermore, for each sample, we only exploit the RGB images and the corresponding 2D and 3D annotations.

The STB dataset contains sequences with six different backgrounds, and each background has two sequences (1500 frame per-sequence). Similar to \cite{hand_geliuhao}, we shift the root joint of STB to make it consistent with RHD, and we split the dataset into training and testing part following the rule given by \cite{hand_zb}. Besides, we only exploit the RGB image and corresponding 3D annotations.

The Dexter dataset consists of six sequences with 2 actors (1 female), and various interactions with a simple object shape. Fingertip positions and cuboid corners are manually annotated for all sequences. Due to the incomplete hand annotation, therefore, similar to \cite{hand_zb}, we use this dataset for evaluating the cross-dataset generalization performance of our framework.

\begin{figure}
\centering
\includegraphics[width=0.45\textwidth]{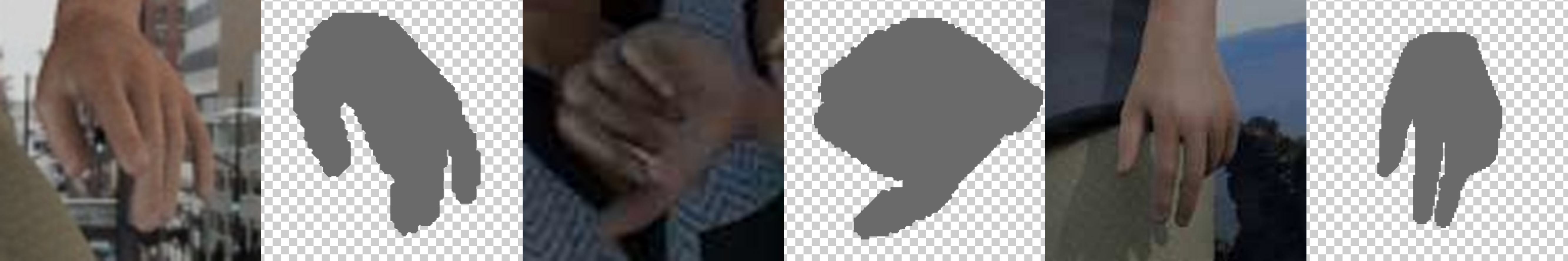}
\caption{{\bf Paired Mask Samples.} Each sample contains two columns. The first column shows the original RGB image; the second column presents the corresponding mask.}
\label{seg_mask}
\end{figure}

We shall point out that, for RHD dataset we obtain the hand mask by parsing the rendered depth map to train the network, Fig. \ref{seg_mask} illustrate some paired hand images and corresponding masks.

{\bf{Data Augmentations.}}
Both RHD and STB share the same augmentation algorithm. We crop each image centered on hand and resize it to $256 \times 256$, then apply scaling (0.9 - 1.2), rotation (-/+$60^{\circ}$).
Finally, to resist the color variety, we perform color-jittering with the following configurations: brightness (0.2 - 1.8), contrast (0.3 - 1.7), saturation (0.4 - 1.6), hue (0.2 - 1.8), and add random noise governed by Gaussian distribution $~\mathcal{N}(0, 0.01)$. 

{\bf{Network design.}}
We use a two-stack Stacked Hourglass as the pose module, the hourglass module output 64 heat-maps which is numerically larger than the number of hand joints (21) for providing more features,  the resolution of heat-maps is $64 \times 64$ and the $\sigma^2$ used for generating $\Phi^{ht}$ is set as 2.5. The iterative regression block consists of a fully convolutional encoder and a fully connected regressor. The encoder is built with 6 stride convolutional layers, for each convolutional layer, the kernel size is $3 \times 3$, and the stride is 2. The regressor is composed of 2 fully connected layers with 2048 neurons, followed by a final layer of 22 neurons. After each fully-connected layer, we insert a dropout layer with drop probability of 0.4 to prevent overfitting.

\begin{figure*}[h]
\centering
\includegraphics[width=1.0\textwidth]{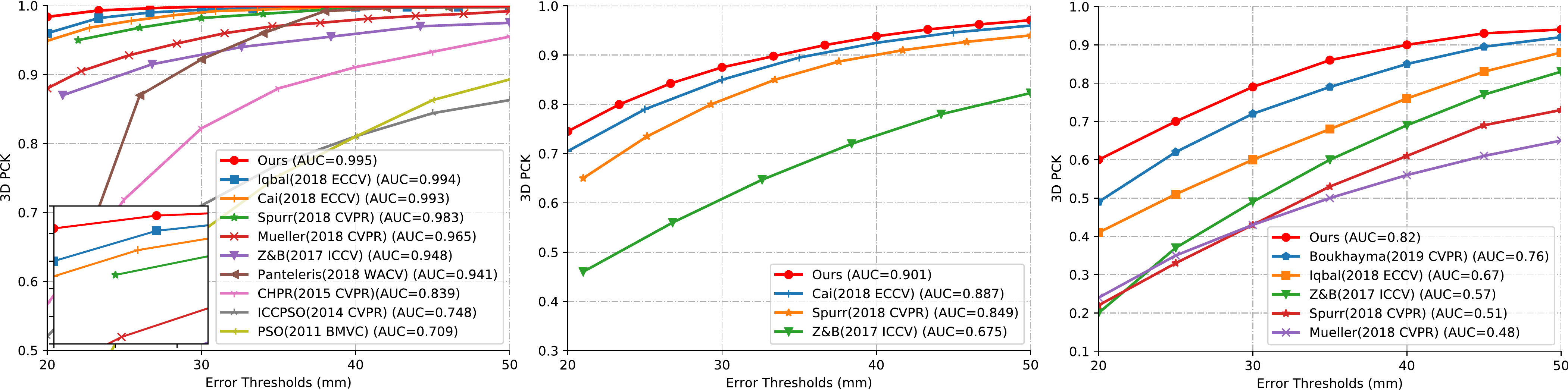}
\caption{{\bf Quantitative Evaluation.}  The left plot presents the 3D PCK performance on the STB dataset,
in which our approach is comparable with state-of-the-art methods. The middle plot shows the 3D PCK performance on the RHD dataset and our method achieves the best results compared with state-of-the-art methods. The right plot demonstrate the 3D PCK performance on the Dexter dataset, our framework outperform state-of-the-art methods by a large margin.}
\label{RHD_STB_eval}
\end{figure*}

{\bf{Training.}}
We implement our framework with Pytorch \cite{paszkepytorch}, and we adopt the end-to-end training manner.
We minimize the total loss in Eq.~\ref{total_loss} to train the whole model, where the value of hyper-parameters $\lambda_{3D}$, $\lambda_{2D}$, $\lambda_{geo}$, $\lambda_{cam}$, $\lambda_{ht}$, $\lambda_{seg}$ are empirically set to 1000, 1, 1, 0.1, 100 and 10 respectively, while one can also tune these hyper-parameters via the grid search method \cite{bergstra2012random}. The RMSprop \cite{tieleman2012lecture} is used to optimize the framework.
We begin training the whole framework with a learning rate $2 \times 10^{-4}$, then decays the learning rate by a factor of $0.7$ when validation loss plateaus.
All the experiments are conducted on GeForce GTX TITAN Xp GPU with CUDA 8.0.

\subsection{Visualization of Appealing Mesh}
In order to verify the quality of the generated mesh and the robustness of the framework in various cases, we draw some representative samples from STB and RHD testing sets as illustrated in Fig.~\ref{appealing_mesh}.
Experiments demonstrate that our framework is capable of generating high-quality, appealing hand meshes.

Further, our framework is robust to reconstruct hand meshes accurately even in severe hard scenarios such as poor lighting, image truncation, self-occlusion, and exaggerated articulation.
For illustration, we refer $S_i$ to the $i$th sample drawn from STB in Fig. \ref{appealing_mesh}, which is similar to $R_i$ for RHD.
The comparison among $R_1$, $R_4$ and $R_5$ reveals that our method can reconstruct the mesh correctly even with poor lighting. 
Beside, $R_2$ indicates our approach can estimate the mesh reasonably even in the presence of image truncation. 
By comparing $S_1$ to $S_5$ together with $R_3$, one can conclude that our framework is capable of recovering anthropometrically reasonable mesh representation even with heavily self-occlusion or complex exaggerated hand pose.

\subsection{Quantitative Evaluation}
\label{Quantitative_Evaluation}
To be consistent with \cite{hand_geliuhao,hand_muller,spurr2018cross,hand_zb}, we evaluate the performance of 3D hand pose estimation with the measure of the area under the curve (AUC) on the percentage of correct key points (PCK) score with different thresholds. 
All the comparison methods are evaluated on both RHD, STB and Dexter datasets, experimental results are shown by Fig. \ref{RHD_STB_eval}. 


On STB dataset, the 3D PCK curves of \cite{hand_geliuhao,iqbal2018hand,hand_muller} are intertwined to each other for the reason that STB dataset is relatively small, and lacks diversity. 
Our method performs competitively with all the comparison methods \cite{hand_geliuhao,iqbal2018hand,hand_muller,panteleris2018using,spurr2018cross,zhang20163d,hand_zb}, which is reasonable considering the saturated performance on this dataset.
In contrast, RHD dataset is relatively complex and more diverse, our method is superior to those proposed by \cite{hand_geliuhao,spurr2018cross,hand_zb} and achieves state-of-the-art result. 
Similarity, on Dexter Object dataset, our method largely outperform most state-of-the art methods \cite{boukhayma20193d,iqbal2018hand,hand_muller,spurr2018cross,hand_zb}.
It is consistent with our expectation that introducing the parametric hand model can greatly help solving the perspective ambiguity issue which in turn improves the accuracy of hand pose estimation.

\begin{figure}
\centering
\includegraphics[width=0.45\textwidth]{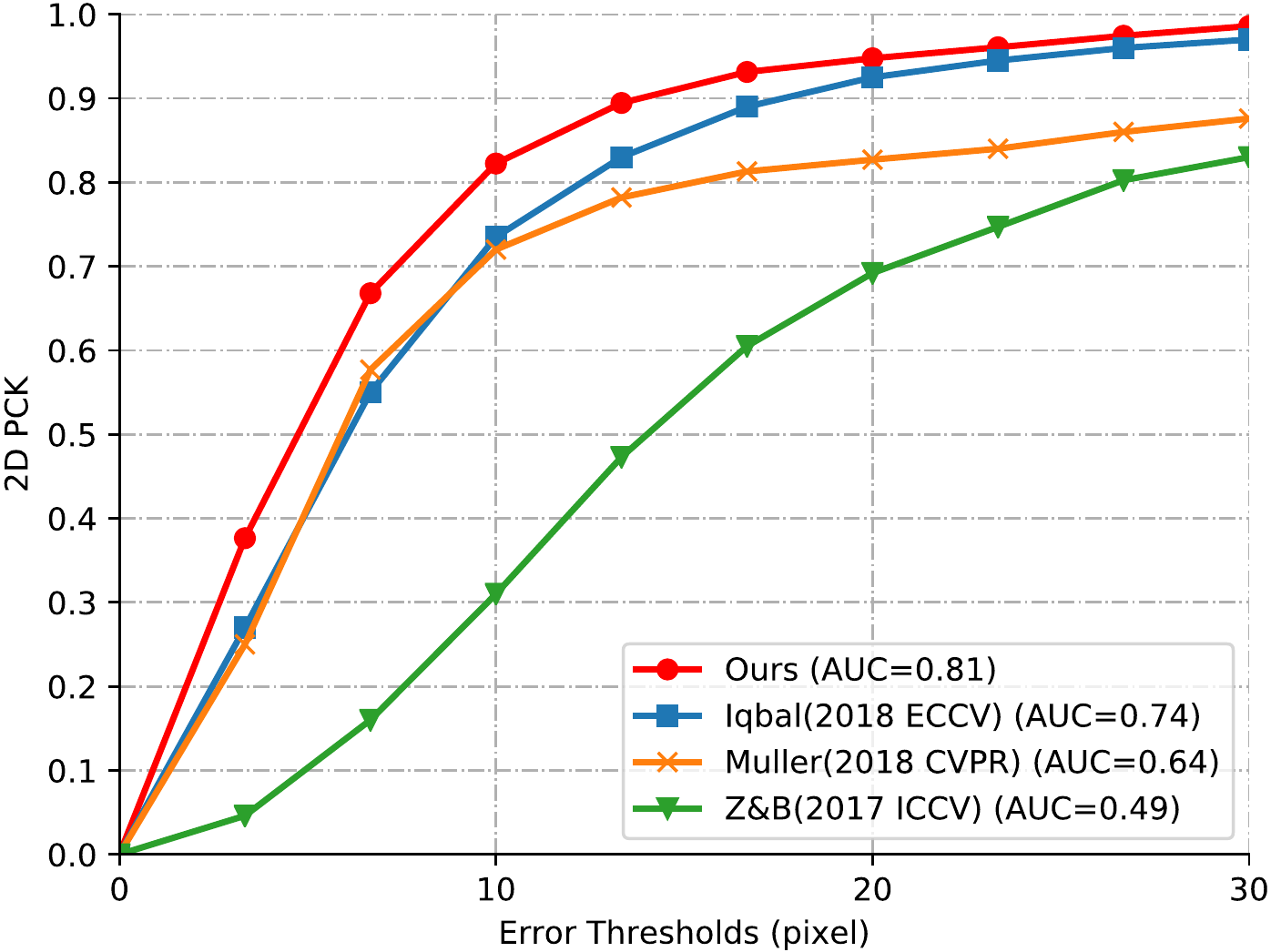}
\caption{{\bf Generalization Performance Evaluation.} The plot illustrates the results of 2D hand pose estimation on the Dexter Object dataset. Our method outperforms those proposed in \cite{iqbal2018hand,hand_muller,hand_zb}.}
\label{dextor}
\end{figure}

\begin{figure*}
\centering
\includegraphics[width=0.95\textwidth]{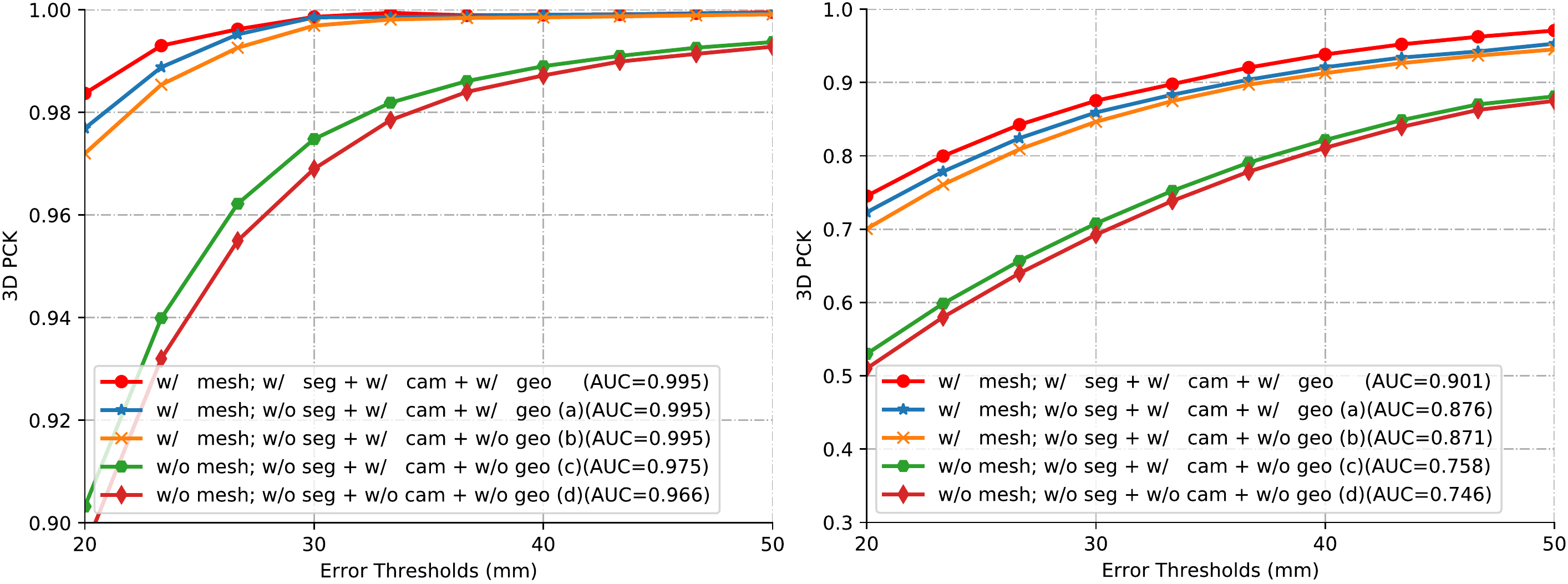}
\caption{{\bf Ablation Studies.} The left plot shows the 3D PCK performance on the STB dataset and the right plot on the RHD dataset.
Note that {\bf w/} and {\bf w/o} abbreviates {\it with} and {\it without} respectively. Besides, {\bf w/ mesh} means obtaining the 3D hand pose from the estimated mesh, {\bf w/o mesh} denotes estimating the 3D joint locations with the iterative regression module, {\bf w/ cam} applies supervision over the pose camera parameters, {\bf w/ geo} denotes taking the geometric loss into consideration, {\bf w/ seg} indicates considering the silhouette consistent constraint.}
\label{RHD_STB_ab}
\end{figure*} 

\begin{figure}
\centering
\includegraphics[width=0.45\textwidth]{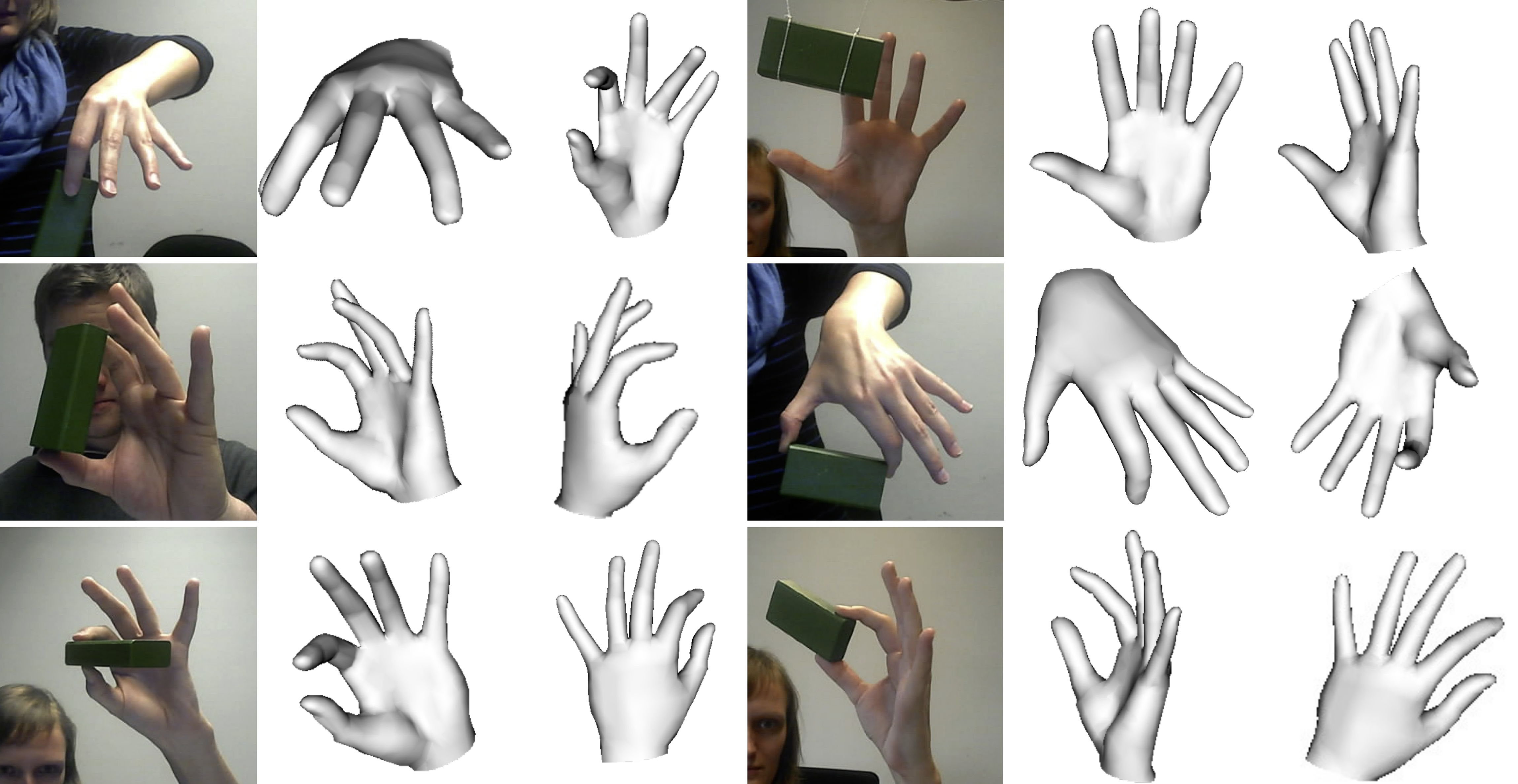}
\caption{{\bf Recovered Meshes from Dexter Dataset.} Each row presents two examples. For each example, the first column shows the RGB image, the second to third columns illustrate the recovered mesh from different views.}
\label{dextor_mesh}
\end{figure}

\subsection{Generalization Performance Evaluation}
Similar to \cite{ iqbal2018hand,hand_muller,hand_zb},  we evaluate the generalization performance of our framework on Dexter dataset \cite{hand_dextor}. Specifically, we train the model on RHD and STB datasets, then perform cross-dataset testing in two ways without fine-tuning.
Firstly, following conventional practice, we take the AUC of 2D PCK as evaluation metric to evaluate the performance on 2D hand pose estimation.
As shown in Fig.~\ref{dextor}, our method achieves better result than those reported in \cite {iqbal2018hand,hand_muller,hand_zb} where the Dexter dataset contains certain  ratio of occluded hand images.
It is not astonishing that the mesh is capable of modeling inherent geometric prior of hand, which further solve uncertainty caused by occlusion. 

Secondly, we examine the quality of estimated hand mesh. As illustrated by Fig. \ref{dextor_mesh}, our framework is capable of reconstructing high-quality mesh even the finger is occluded by some foreign matter or in the presence of high dynamic range lighting. The above experiments verified the generalization performance of our framework, proved the usefulness of mesh in solving the ambiguity caused by occlusion, and implied the essential importance of introducing the parametric model for hand pose estimation.

\subsection{Ablation Studies}

To better understand the impact of different design choices, we evaluate our framework in various settings. Four main design choices exist in this work: the effect of introducing parametric model on the task of pose estimation, the effect of the silhouette consistent constraint, the impact of pose supervision over camera parameters, and the influence of the geometric loss. We still take AUC of 3D PCK as the evaluation metric for RHD and STB datasets.
Fig. \ref{RHD_STB_ab} compares our full scheme {\bf(w/ mesh + w/ cam + w/ geo + w/ seg)} with four alternatives, experiments a , b, c and d, under various settings.

We find that  introducing parametric mesh model improve the performance of hand pose significantly, since the mesh model captures comprehensive properties of hand.
Besides, adding specific geometric constraints over fingers yield marginal performance improvement because the mesh model has already covered the inherent geometric constraints of hand.  
Furthermore, pose  supervision over camera parameters gained certain accuracy improvement because the $\mathcal{L}_{cam}$ term plays the role of calibrating the projection procedure from 3D to 2D. 
Equal important, penalize misalignment between the rendered mask and the ground-truth silhouette also gained performance improvement since the $\mathcal{L}_{seg}$ term can refine the hand shape and pose prediction.


\section{Conclusion}
\label{sec:Conclusion}

Hand image understanding is a widely researched topic and has many real-world vision applications.
While hand pose estimation has been well studied in the literature, there still exists little research on the problem of hand mesh recovery from single RGB images.
The proposed HAMR enriches this field by leveraging from a generic 3D hand model to achieve a mesh representation.
With the mesh model, the framework can produce 2D and 3D hand pose estimations as well.
Qualitatively, HAMR successfully recovered reasonable hand meshes from single RGB images even under severe occlusions.
Quantitatively, the superiority of HAMR was empirically confirmed on 2D and 3D hand pose estimation tasks compared with state-of-the-art methods.
Along this direction, one can expect future advancements in both hand mesh recovery and pose estimation.

\section{Supplementary Material}
We have exploited the segmentation mask to train the network, regard this concern, this additional document quantitative analysis the accuracy of hand segmentation, by projecting the reconstructed hand mesh, which also reflects the precision of the reconstructed mesh from one aspect.
Following conventional, we use the mIoU (mean Intersection over Union) as the evaluation metric.
We shall point out here, since we can not deduce the ground-truth mask on the STB and Dexter datasets, as a result, we conduct the following experiment on RHD dataset.

\begin{figure}[hbp]
\includegraphics[width=1.0 \columnwidth]{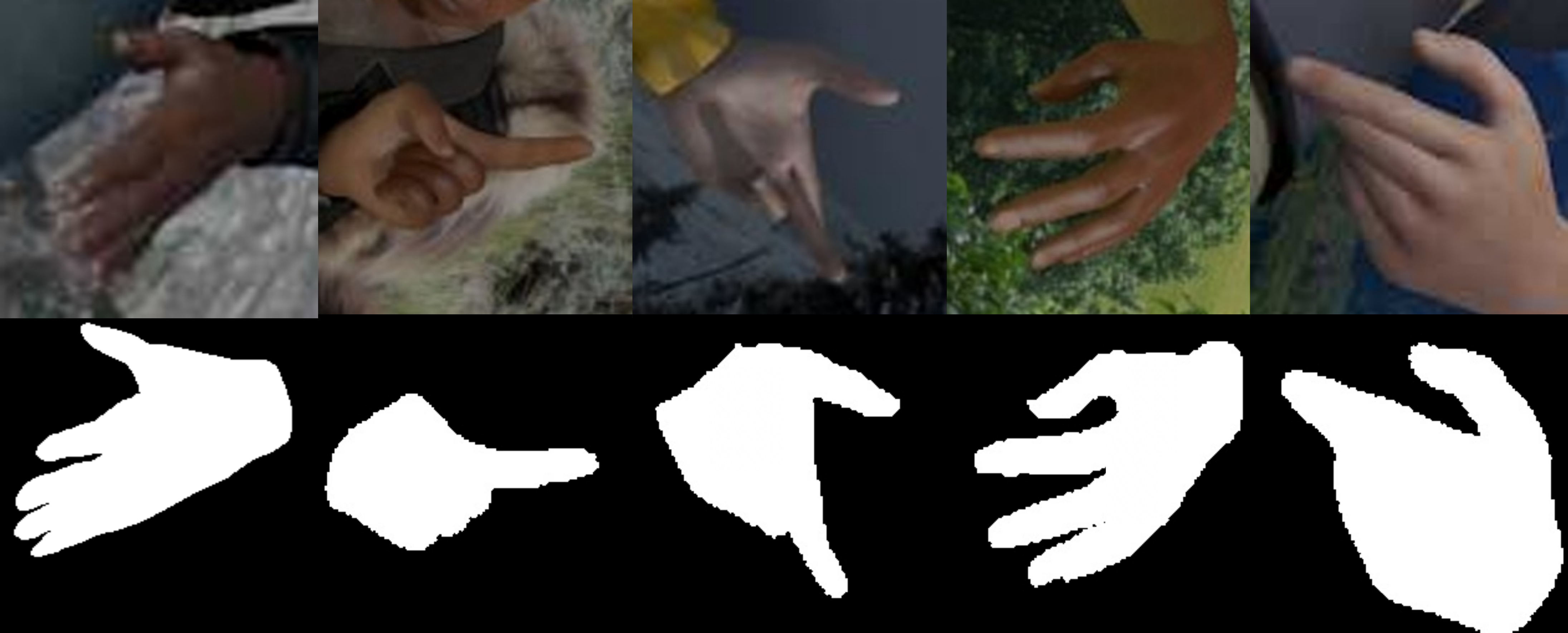}
\caption{{\bf Segmentation Examples.} The graph comprising two rows, the first row presents the RGB images drawn from the RHD testing part, and the second row demonstrates the segmentation mask of each example.}
\label{Fig_Mask}
\end{figure}

The Fig. \ref{Fig_Mask} presents some examples of segmentation mask and Tab. \ref{tab_tab} gives the quantitative comparison result.

\begin{table}[h!]
  \begin{center}
    \caption{Performance of segmentation on RHD dataset.}
    \label{tab_tab}
    \begin{tabular}{l|cr} 
      \textbf{Method} & \textbf{mIoU}\\
      \hline
      HMR \cite{kanazawa2018end} & 0.750\\
      BodyNet \cite{varol2018bodynet} & 0.852 \\
      SMPLify \cite{bogo2016keep} & 0.739\\
      Georgios {\emph {et.al}} \cite{pavlakos2018learning} ~~~~~~~~~~~~~~~~~~& ~~~0.806~~~\\
      DeepLab \cite{chen2018deeplab} & 0.924\\
      \hline 
      HAMR (Ours) & 0.931\\
      \hline 
    \end{tabular}
  \end{center}
\end{table}

Quantitative experimental result illustrated in Tab. \ref{tab_tab} reveals that our methods outperforms all those state-of-the art methods. We are not astonished by the outstanding performance, since the introducing of parametric hand model can solve the ambiguity, in addition the silhouette consistent loss can refine the hand shape and pose prediction.

\newpage
{\small
\bibliographystyle{ieee_fullname}
\bibliography{reference.bib}
}

\end{document}